\documentclass[10pt,twocolumn,letterpaper]{article}

\usepackage[pagenumbers]{iccv} 

\definecolor{iccvblue}{rgb}{0.21,0.49,0.74}
\usepackage[pagebackref,breaklinks,colorlinks,allcolors=iccvblue]{hyperref}

\usepackage[accsupp]{axessibility}  

\usepackage{amsthm}

\usepackage{algorithm}
\usepackage{algpseudocode}

\usepackage{tikz}
\usetikzlibrary{tikzmark}
\usetikzlibrary{backgrounds}
\usepackage{cancel}
\usepackage{accents}
\usepackage{amsmath}
\usepackage{arydshln}
\usepackage{pifont}
\usepackage{pgfplots}
\usepackage{pgfplotstable}

\usepackage{tabularray}
\usepackage{multirow}
\usepackage{fancybox}
\usepackage[outline]{contour}
\usepackage{color, colortbl}

\pgfplotsset{compat=newest}
\definecolor{color1}{RGB}{196, 164, 132} 
\definecolor{color2}{RGB}{30,144,255} 
\definecolor{color3}{RGB}{255, 16, 240} 

\def\paperID{8798} 
\def\confName{ICCV}
\def\confYear{2025}

\title{Self-Calibrated Variance-Stabilizing Transformations for Real-World Image Denoising}

\author{Sébastien Herbreteau$^{1, 2, 3}$ and Michael Unser$^{2}$  \\
$^{1}$CIBM Center for Biomedical Imaging, Switzerland\\
$^{2}$Biomedical Imaging Group, École polytechnique fédérale de Lausanne (EPFL),  Switzerland\\
$^{3}$Univ Rennes, Ensai, CNRS, CREST—UMR 9194, F-35000 Rennes, France \\
{\tt\small sebastien.herbreteau@ensai.fr \quad michael.unser@epfl.ch}
}

\begin{document}
\maketitle

\begin{abstract}

Supervised deep learning has become the method of choice for image denoising. It involves the training of neural networks on large datasets composed of  pairs of noisy and clean images. However, the necessity of training data that are specific to the targeted application constrains the widespread use of denoising networks. Recently, several approaches have been developed to overcome this difficulty by whether artificially generating realistic clean/noisy image pairs, or training exclusively on noisy images. In this paper, we show that, contrary to popular belief, denoising networks specialized in the removal of Gaussian noise can be efficiently leveraged in favor of real-world image denoising, even without additional training. For this to happen, an appropriate variance-stabilizing transform (VST) has to be applied beforehand. We propose an algorithm termed Noise2VST for the learning of such a model-free VST. Our approach requires only the input noisy image and an off-the-shelf Gaussian denoiser. We demonstrate through extensive experiments the efficiency and superiority of  Noise2VST  in comparison to existing methods trained in the absence of specific clean/noisy pairs.

\end{abstract} 
\section{Introduction}

Image denoising involves the retrieval of a latent clean image $\boldsymbol{s} 
\in \mathbb{R}^N$ from its noisy observation $\boldsymbol{z} \in \mathbb{R}^N$. 
In its most studied formulation, each noisy pixel $z_i$ is assumed to independently follow an $s_i$-centered Gaussian distribution with variance $\sigma^2$, in accordance with
\begin{equation}
    z_i \sim  \mathcal{N}(s_i, \sigma^2) \,.
\end{equation}
\noindent  The emergence of machine learning techniques has enabled rapid and significant progress in the removal of Gaussian  noise \cite{dncnn, ffdnet, drunet, mwcnn, restormer, nlrn, xformer, swinir}. 
The most successful denoising techniques rely on datasets composed of numerous pairs  consisting of a  clean image and its noisy counterpart, synthesized by artificially adding Gaussian noise to clean images. Such training datasets are particularly easy to build, given  the wide availability of images that one can assume to be noise-free, along with the simple nature of the simulated noise.

Nevertheless, empirical observations tend to indicate that models trained specifically to remove  Gaussian noise  do not generalize well to real-world scenarios \cite{DND, SIDD, zhang2019poisson, w2s}. Since real noise stems from diverse sources of degradation during data acquisition, it is unlikely to follow a normal distribution. While recommended practice involves the training of a specialized model on  pairs of clean/noisy images from the targeted application \cite{DND, SIDD, zhang2019poisson, w2s}, the assembly of a sufficiently large, high-quality dataset can be exceedingly time-consuming or even impractical in certain contexts. This explains  the growing focus on methods that train without specific clean/noisy image pairs, whether it is by generating realistic clean/noisy image pairs artificially \cite{Ryou_2024_CVPR, unprocessing, cycleisp, PNGAN, scunet}, training solely on noisy images \cite{N2V, N2S, noise2noise, neighbor2neighbor, blind2unblind}, or even without relying on any dataset \cite{ZS-N2N, S2S, N2F, DIP}.

In this paper, we show that  deep neural networks specialized in the removal of Gaussian noise, specifically,  can also be efficiently harnessed to tackle the denoising of real-world images. Our work draws inspiration from the traditional theory of variance-stabilizing transformations (VSTs), for which the most popular representative is the Anscombe transform \cite{anscombe_poisson}, and revisits it with a dose of machine learning. We propose indeed to learn a function $f_{\boldsymbol{\theta}} : \mathbb{R} \mapsto \mathbb{R}$ that we apply pixel-wise to the input image $\boldsymbol{z}$ and that is such that the resulting transformed  distribution of the noise adjusts approximately to a Gaussian distribution. More precisely, $f_{\boldsymbol{\theta}}(\boldsymbol{z})$ is intended to be processed by an off-the-shelf Gaussian denoising network, with frozen weights, before returning to the original domain by ways of the inverse transformation. Contrary to traditional VST approaches, we do not assume any model for the noise, except that it is zero-mean and spatially independent. The training of $f_{\boldsymbol{\theta}}$ uses the recent blind-spot strategy introduced in \cite{N2V, N2S}, applied here on a single noisy image and with guaranties regarding overfitting due to the few parameters being optimized.  Our extensive experiments demonstrate that our proposed zero-shot denoising framework provides reconstructions of remarkable quality.

The contributions of our work are as follows:
\begin{enumerate}
\item  The design of  a  model-free VST under the form of an increasing continuous piecewise linear function. 
    \item The proposal of a novel framework to tackle real-world noise removal, requiring only an input noisy image and an off-the-shelf Gaussian denoiser. 
    \item The demonstration of the superior performance of our approach  compared
with state-of-the-art methods trained without specific ground truths, at a limited computational cost.
\end{enumerate}

\section{Related Work}

Denoising methods without specific clean/noisy image pairs can be classified into two main categories, depending on whether an external dataset is used or not.

\subsection{Dataset-Based Methods}


\paragraph{Nonspecific Clean/Noisy Pairs} FFDNet \cite{ffdnet} learns to map images corrupted with artificial additive white Gaussian noise to their latent clean representation. Interestingly, this network, which requires an estimation of the noise level as additional input, demonstrates superior generalization to real-world noise compared to its blind counterpart DnCNN \cite{dncnn}.
Several methods \cite{Ryou_2024_CVPR, unprocessing, cycleisp, PNGAN, scunet} improve the synthetic clean/noisy image-pair generation by carefully considering the noise properties of camera sensors \cite{DND, SIDD}. Finally, the training of denoising networks from an unpaired set of clean and noisy images is investigated in  \cite{dbsn}.

\paragraph{Specific Noisy/Noisy pairs}

Noise2Noise \cite{noise2noise} assumes that, for the same underlying clean image, two independent noisy observations are available. The authors  demonstrate that a training on these noisy pairs is equivalent to a training in a supervised setting with specific clean/noisy images.  Building upon this idea, the blind-spot approach is introduced in \cite{N2V, N2S}. It prevents neural networks from learning the identity, thereby enabling training on individual noisy images. However, blind-spot networks achieve subpar performance in practice because they intentionally avoid using information from the pixel being denoised and solely rely on data from their neighborhood. The blind-spot estimation is improved in \cite{laine, pN2V} by incorporating this missing information at inference but  a model for the noise is necessary, limiting their use
in practice. Finally, several methods \cite{noisier2noise, R2R, zou2023iterative, neighbor2neighbor, blind2unblind}  generate training pairs in the spirit of  Noise2Noise \cite{noise2noise} from individual noisy images, by employing subsampling techniques or by adding artificial supplementary noise.


\subsection{Dataset-Free Methods}

\paragraph{Training on the Noisy Input Image}

The implicit regularization effect of a convolutional network is used in \cite{DIP} for training directly on the noisy input image, preventing the network from overfitting to the noise with early stopping. 
S2S \cite{S2S} learns on  pairs
of Bernoulli-sampled instances of the input image, leveraging dropout \cite{dropout} as regularization technique. Finally, the recent N2F \cite{N2F} and ZS-N2N \cite{ZS-N2N} methods adapt the subsampling strategy of \cite{neighbor2neighbor} in a dataset-free context for practical zero-shot denoising with a shallow network.


\paragraph{Model-Based Methods} A majority of non-learning methods \cite{BM3D, nlridge, WNNM, nlbayes, nlmeans, ksvd, EPLL_unsupervised} are built upon the assumption that the noise in images is white, additive, and Gaussian. 
These methods  are generally combined with a variance-stabilizing transformation such as the Anscombe transform \cite{anscombe_poisson, GAT2} to tackle real-world noise removal. This  strategy is adopted in NoiseClinic  \cite{noise_clinic}, for which the denoiser at the core is a multiscale version of NL-Bayes \cite{nlbayes}. Other approaches \cite{TWSC, purelet} try to estimate the clean image directly by dispensing with the usual Gaussian setting.

\section{Method}

In this section, we first recall the traditional theory behind VSTs, with a focus on Poisson-Gaussian noise. Then, we show how VSTs and their optimal inverses can be modeled using splines. Finally, we present a self-supervised strategy for learning an optimal VST using only the input noisy image $\boldsymbol{z}$.

\subsection{Variance-Stabilizing Transformations}

\input{figure_method2}

VSTs are powerful tools in the realm of statistics and data analysis \cite{GAT2}. 
In denoising contexts where the noise variance is pixel-dependent, their purpose is to modify the input noisy image  to achieve homoscedasticity, therefore constraining the noise variance to remain approximately constant across all pixel intensities.

\paragraph{Focus on Poisson-Gaussian Noise} VSTs were particularly studied in the case of  mixed Poisson-Gaussian noise, the most common model for real-world image noise, modeling both the shot and the read noise \cite{photon_noise, DND, unprocessing, cycleisp, poisson_estimation}. The noise model reads independently for each pixel as
\begin{equation}
    z_i \sim a \mathcal{P}(s_i/a) + \mathcal{N}(0, b )\,,
    \label{pg_model}
\end{equation}
\noindent where $a, b \in (0, \infty)$ are the noise parameters. Note that the noise variance does depend on the pixel intensities since $\sigma^2(z_i) = a s_i+ b$. In order to process such a type of noise, the generalized Anscombe transform (GAT) \cite{GAT2} is often employed as  VST, with
\begin{equation}
    f_{\text{GAT}}: z \mapsto \frac{2}{a} \sqrt{\max\left(az + \frac{3}{8} a^2 + b, 0\right)}\,.
    \label{GAT}
\end{equation}

\noindent This mapping is designed in such a way that the noise in $f_{\text{GAT}}(\boldsymbol{z})$ is approximately homoscedastic Gaussian with unit variance. A Gaussian denoiser $D$ can then be leveraged to process $f_{\text{GAT}}(\boldsymbol{z})$. One then  returns to the original range by applying the inverse transformation $f_{\text{GAT}}^{\operatorname{inv}}$. This results in
\begin{equation}
    \hat{\boldsymbol{s}} = (f_{\text{GAT}}^{\operatorname{inv}} \circ D \circ f_{\text{GAT}})(\boldsymbol{z}) \,,
\end{equation}
\noindent 
where $\hat{\boldsymbol{s}}$ denotes the final estimate of the noise-free image. Note that the optimal  ``inverse'' transformation $f_{\text{GAT}}^{\operatorname{inv}}$ is in general slightly different from the algebraic inverse $f_{\text{GAT}}^{-1}$ since the latter tends to introduce an undesired bias \cite{anscombe_inverse}. Instead, it is recommended \cite{anscombe_inverse} to take $f_{\text{GAT}}^{\operatorname{inv}}$ as the exact unbiased inverse given by the implicit mapping $\mathbb{E}[f_{\text{GAT}}(z) | s] \mapsto s$, where $z$ denotes the $s$-mean real-valued random variable that follows \eqref{pg_model}. Approximate closed-from expressions of the unbiased inverse for Poisson-Gaussian distributions  are proposed in \cite{anscombe_inverse}.

While theoretically powerful, the VST approach has a major drawback in practice: it is model-dependent. Indeed, it is important to remember that models are always mere approximations of reality. Some recent works \cite{scunet, cycleisp} suggest that real image noise is too complex to be modeled with standard distributions. Moreover, even when the parametric probability
distribution is known, the estimation of the true parameters of the noise distribution (parameters $a$ and $b$ in the case of Poisson-Gaussian noise) may be challenging \cite{poisson_estimation, poisson_estimation2}. A poor estimation of these can obviously result in a severe drop in denoising performance. Finally,  the derivation of a VST for a given noise distribution often involves a combination of statistical theory, empirical analysis, and domain expertise to identify the most appropriate transformation, making its application tedious in many cases. Let us mention here the pioneering works of \cite{anscombe_poisson, freeman1950transformations, veevers1971variance, bartlett1936square, curtiss1943transformations, tibshirani1988estimating} aimed at achieving the stabilization of usual probability distributions in the asymptotic sense. We also mention the recent works of \cite{FBI, VST-Net} that combine deep learning and VSTs.

\subsection{VST  Modeling with Splines}

Our goal is to construct a function $f : \mathbb{R} \mapsto \mathbb{R}$  that emulates a VST, as well as its inverse transform $f^{\operatorname{inv}}$. We propose to search for the optimal transforms among the family of increasing continuous piecewise linear (CPWL) mappings,  with finitely many pieces. This choice is justified by the fact that splines are universal approximators \cite{spline1, spline2}.
Moreover, considering monotonically increasing functions is natural in order to preserve the  order of pixels in the transform domain. 

An increasing CPWL function $f$ is completely parameterized by the knots $\{(x_i, y_i) \}_{1\leq i \leq n}$ between which $f$ exhibits a linear behavior, where $ \boldsymbol{x} = (x_i)$  and $\boldsymbol{y} = (y_i) = (f(x_i))$ are  increasing finite sequences of $\mathbb{R}$. More precisely, we have that
\begin{equation}
f: z \mapsto  \frac{  y_{\iota(z; \boldsymbol{x})  + 1} - y_{\iota(z; \boldsymbol{x})}}{x_{\iota(z; \boldsymbol{x}) + 1} -x_{\iota(z; \boldsymbol{x})}}  (z-x_{\iota(z; \boldsymbol{x})}) + y_{\iota(z; \boldsymbol{x})}\,,
\label{eq:forward}
\end{equation}
where
$\iota(z; \boldsymbol{x}) = \max  \left\{ i \in \{1, \ldots, n-1\} \; | \;  x_i \leq z \right\} \cup \{1\}$. An example of a such a function is shown in Figure \ref{method}d (blue solid line). Instead of searching $f$ among the entire parametric family of  increasing CPWL functions, we propose to restrict the search space by arbitrarily fixing the $x_i$ to a uniform spacing over $[\boldsymbol{z}_{\operatorname{min}}, \boldsymbol{z}_{\operatorname{max}}]$ such that
\begin{equation}
  x_i =  (\boldsymbol{z}_{\operatorname{max}}-\boldsymbol{z}_{\operatorname{min}})  \frac{i-1}{n-1} + \boldsymbol{z}_{\operatorname{min}}\,, 
  \label{eq:xi}
\end{equation}
where $\boldsymbol{z}_{\operatorname{min}}$ and $\boldsymbol{z}_{\operatorname{max}}$ denote the minimum and maximum pixel values of $\boldsymbol{z}$, respectively. Therefore,  only the values of the $y_i$  are parameterized with a vector $\boldsymbol{\theta} \in \mathbb{R}^n$ such that
\begin{equation}
 y_i = \theta_1 + \sum_{j=2}^{i}  \exp(\theta_j)\,. 
 \label{parametrization}
\end{equation}

\noindent In the Supplementary Material, we show that $\theta_1$ can actually be fixed to $0$ without loss of generality. This parameterization guarantees that $y_i < y_{i+1}$.  The  resulting parameterized function with \eqref{parametrization} is referred to as $f_{\boldsymbol{\theta}}$. 

Regarding the inverse transformation, we propose to constraint it to be close to the algebraic inverse of $f_{\boldsymbol{\theta}}$, while permitting 
a certain degree of freedom, according to
\begin{equation}
f^{\operatorname{inv}}_{\boldsymbol{\theta}, \alpha, \beta}: z \mapsto f_{\boldsymbol{\theta}}^{-1}(z) + \alpha z + \beta\,.
\label{eq:inverse}
\end{equation}
The affine function $z\mapsto \alpha z + \beta$ is added with the hope that it can mitigate the undesired bias usually introduced by the algebraic inverse $f_{\boldsymbol{\theta}}^{-1}$ alone (see Supplementary Material for an ablation study on the design of the inverse transform). Note that $f_{\boldsymbol{\theta}}^{-1}$ always exists and is an increasing CPWL function since $f_{\boldsymbol{\theta}}$ is. More precisely, we have that
\begin{equation}
f^{-1}_{\boldsymbol{\theta}}: z \mapsto  \frac{x_{\iota(z; \boldsymbol{y})  + 1} - x_{\iota(z; \boldsymbol{y})}}{y_{\iota(z; \boldsymbol{y}) + 1}-y_{\iota(z; \boldsymbol{y})}} ( z-y_{\iota(z; \boldsymbol{y})}) + x_{\iota(z; \boldsymbol{y})}\,,
\end{equation}
where the $y_i$ implicitly depend on $\boldsymbol{\theta}$ through \eqref{parametrization}. Note that $f_{\boldsymbol{\theta}}$ and $f^{\operatorname{inv}}_{\boldsymbol{\theta}, \alpha, \beta}$ do share the same parameters $\boldsymbol{\theta}$, the inverse transform having only  two extra parameters $\alpha$ and $\beta$.

All in all, the total number of learnable parameters for the forward and backward transformations is $n+2$. In practice, we take $n=128$.

\subsection{Training with a Blind-Spot Denoiser}

Equipped with the two real-valued parametric functions $f_{\boldsymbol{\theta}}$ and $f^{\operatorname{inv}}_{\boldsymbol{\theta}, \alpha, \beta}$, and an off-the-shelf differentiable Gaussian denoiser $D$ (with frozen weights if $D$ is a neural network), our goal is to learn the optimal parameters $\boldsymbol{\theta}, \alpha$, and $\beta$ that best remove noise from the input image $\boldsymbol{z}$. In other words, we aim to minimize the usual squared $\ell_2$ loss 
\begin{equation}
     \mathcal{L}^D_{\boldsymbol{\theta}, \alpha, \beta}(\boldsymbol{z}, \boldsymbol{s})=  \| (f^{\operatorname{inv}}_{\boldsymbol{\theta}, \alpha, \beta} \circ D \circ f_{\boldsymbol{\theta}})(\boldsymbol{z}) - \boldsymbol{s}  \|_2^2 \,.
    \label{ideal_strategy}
\end{equation}
Of course, the noise-free image $\boldsymbol{s}$ is unknown, so  that \eqref{ideal_strategy} cannot be computed in practice. However,  there is one setting where the  minimization of  \eqref{ideal_strategy} can still be achieved: when the denoiser is blind-spot  \cite{N2S, N2V}.

\paragraph{Blind-Spot Denoisers} A denoiser is termed blind-spot, denoted by a bar,  if $[\bar{D}(\boldsymbol{z})]_i$ does not depend on $z_i$.  The raison d'être of blind-spot functions is that, under the assumption that $\boldsymbol{z}$ \textit{is an unbiased estimator of} $\boldsymbol{s}$ \textit{with spatially uncorrelated noise}, the usual quadratic risk and the self-supervised one are equal up to a constant, which is expressed as
\begin{equation}
    \mathbb{E} \| \bar{D}(\boldsymbol{z}) - \boldsymbol{z} \|_2^2 = \mathbb{E} \| \bar{D}(\boldsymbol{z}) - \boldsymbol{s} \|_2^2 + \operatorname{const}. 
\end{equation}

\noindent This remarkable property was exploited by \cite{N2S, N2V} to train denoising networks in the absence of ground truths. Indeed, minimizing the empirical self-supervised loss on large databases of single noisy images ultimately amounts to minimize the usual quadratic loss. This enables a fully self-supervised training approach for image denoising.
\bigskip

The equivalence between the minimization of the self-supervised risk and the classic one for blind-spot functions can also  be advantageously exploited  for VST self-supervised learning. Indeed, since the composition by an element-wise function does preserve the blind-spot property, we have that 
 $f^{\operatorname{inv}}_{\boldsymbol{\theta}, \alpha, \beta} \circ \bar{D} \circ f_{\boldsymbol{\theta}}$ is blind-spot as long as $\bar{D}$ is. Thus, provided that the amount of noisy data is sufficiently large, we have that
 \begin{equation}
\mathcal{L}^{\bar{D}}_{\boldsymbol{\theta}, \alpha, \beta}(\boldsymbol{z}, \boldsymbol{z}) = \mathcal{L}^{\bar{D}}_{\boldsymbol{\theta}, \alpha, \beta}(\boldsymbol{z}, \boldsymbol{s}) + \operatorname{const}.
\label{equa_n2s}
\end{equation}

\noindent Therefore, the search for the parameters $\boldsymbol{\theta}, \alpha, \beta$ that minimize $\mathcal{L}^{\bar{D}}_{\boldsymbol{\theta}, \alpha, \beta}(\boldsymbol{z}, \boldsymbol{s})$  amounts to minimizing $\mathcal{L}^{\bar{D}}_{\boldsymbol{\theta}, \alpha, \beta}(\boldsymbol{z}, \boldsymbol{z})$, a quantity that does not depend on the true signal $\boldsymbol{s}$ anymore. In practice, this strategy still holds without overfitting due to the very few parameters involved---recall that the weights of the off-the-shelf denoiser are frozen. We show in Figure \ref{method}a the proposed self-supervised training procedure for VST learning, while Figure \ref{method}d displays the learned VST. Refer to the Supplementary Material for insights on how  this transformation stabilizes the variance.

In practice, we build $\bar{D}$ from a classic Gaussian denoiser $D$ with no additional training,  according to the method proposed  by \cite{N2S}. This method requires only two components: a partition $\mathcal{J}$ of the pixels of $\boldsymbol{z}$ such that neighboring pixels are in different subsets; and a function $\eta$ replacing each pixel with an average of its neighbors. The blind-spot denoiser is then realized by
\begin{equation} \label{trick}
\forall J \in \mathcal{J},  \quad    \bar{D}(\boldsymbol{z})_J = D(\boldsymbol{1}_J \cdot \eta(\boldsymbol{z}) + \boldsymbol{1}_{J^\mathrm{c}} \cdot \boldsymbol{z})_J\,,
\end{equation}
\noindent where $J^{\mathrm{c}}$ denotes the complement of $J$. We choose $\mathcal{J}$ and $\eta$ as recommended by \cite{N2S}.

\begin{algorithm}[t]
\caption{Noise2VST}
\begin{algorithmic}
\Require Noisy image $\boldsymbol{z}$, off-the-shelf classical and blind-spot Gaussian denoisers $D$ and $\bar{D}$, respectively.
\Ensure Denoised image $\hat{\boldsymbol{s}}$.

\State Set $\alpha, \beta$ to $0$, and $\boldsymbol{\theta}$ such that $f_{\boldsymbol{\theta}}$ is the identity function. 

\While{\textit{not converged}}

\State Update $f_{\boldsymbol{\theta}}$ and $f^{\operatorname{inv}}_{\boldsymbol{\theta}, \alpha, \beta}$  so as to minimize the loss 
\State $\mathcal{L}^{\bar{D}}_{\boldsymbol{\theta}, \alpha, \beta}(\boldsymbol{z}, \boldsymbol{z}) = \|(f^{\operatorname{inv}}_{\boldsymbol{\theta}, \alpha, \beta} \circ \bar{D} \circ f_{\boldsymbol{\theta}})(\boldsymbol{z}) - \boldsymbol{z} \|_{2}^2\,.$ 

\EndWhile 

\State \Return the denoised image $\hat{\boldsymbol{s}} = (f^{\operatorname{inv}}_{\boldsymbol{\theta}, \alpha, \beta} \circ D \circ f_{\boldsymbol{\theta}})(\boldsymbol{z})$.
\end{algorithmic}
\label{algo1}
\end{algorithm}

\subsection{Inference with Visible Blind Spots}

In spite of their great theoretical interest, blind-spot denoisers are unfortunately of limited practical value due to the constraint of deliberately not  using the information of the pixel being denoised. Indeed, except from the parts of the signal that are easily predictable (for example, uniform regions), counting exclusively on the information provided by the neighboring pixels is  suboptimal. For instance, the denoising of a uniformly black image with a single white pixel at its center, as achieved by a blind-spot denoiser, will result in a  central white pixel that will be lost and wrongly replaced by a black one. This is why several methods have been proposed \cite{laine, pN2V} to recover this lost information a posteriori using Bayesian approaches, at the cost of a dependence on   a specific noise model.

In our case, however, a simple approach is possible for enhanced performance: replace the blind-spot denoiser $\bar{D}$ (that served during training) by a classic Gaussian denoiser $D$ at inference. This substitution at inference is legitimate since  \textit{an optimal VST ought not depend on the denoiser}, whether it is blind-spot or not, but only on the noise distribution. This particularity allows us to switch from a blind-spot regime during training to a fully visible one at inference, avoiding in particular the  checkerboard artifacts   inherent to blind-spot denoisers \cite{N2V2} (see Figure \ref{method}a). Note that switching off the blind-spot at inference is a strategy that proved to yield increased performance \cite{blind2unblind, neighbor2neighbor}. We illustrate in Figure \ref{method}c  the proposed inference procedure, which significantly enhances performance.  The final proposed denoising framework, named Noise2VST, is shown in Algorithm \ref{algo1}. Note that, due to the independence of the optimal VST on the denoiser, $\bar{D}$ and $D$ do not need to share similar architectures. In particular, a fast blind-spot denoiser can be used during training, while a more costly and powerful classic one can be leveraged for inference. This fast alternative is denoted Noise2VST$\dagger$ in the sequel.

\section{Experimental Results}

In this section, we first describe the implementation details of the proposed zero-shot method (Algorithm \ref{algo1}). We then evaluate its performance by comparing it with its state-of-the-art counterparts trained without ground truths. 
We used the implementations provided by the authors as well as the corresponding pre-trained weights when available. The PSNR and SSIM values, used for assessment, are sourced from \cite{blind2unblind, yasarla2024self, zou2023iterative, deconoising, pN2V, FBI}, with the exception of ZS-N2N \cite{ZS-N2N}, which we carefully evaluated ourselves. The comparisons include: 1) synthetic Poisson denoising in sRGB space; and 2) real-world noisy image denoising in raw and raw-RGB space. Finally, we conducted running-time comparisons to demonstrate the efficiency of  our method.

\begin{table}[t]
\scriptsize
  \centering
  \setlength{\abovecaptionskip}{0.1cm} 
  \setlength\tabcolsep{4pt} 
  \begin{tabular}[b]{clccc}
    \toprule
       & Method & KODAK & BSD300 & SET14 \\
    \midrule
      \multirow{14}{*}{\shortstack[c]{\rotatebox{90}{Poisson $\lambda\in[5,50]$}}} 
        & Baseline,  N2C + GAT \cite{drunet,{anscombe_inverse}} & 31.63/0.865  & 29.92/0.850  & 30.66/0.854  \\ 
        \cline{2-5}
        & CBM3D + GAT \cite{BM3D, anscombe_inverse} & 29.40/0.836 & 28.22/0.815 & 28.51/0.817 \\ 
        
        & N2V \cite{N2V} &  30.55/0.844 & 29.46/0.844  & 29.44/0.831 \\

        & Nr2N \cite{noisier2noise} &  30.31/0.812 & 29.45/0.821 & 29.40/0.812 \\

        & S2S \cite{S2S} & 29.06/0.834 & 28.15/0.817 & 28.83/0.841 \\

        & DBSN  & 29.60/0.811 & 27.81/0.771 & 28.72/0.800 \\
        

        & SSDN \cite{laine} & 30.88/0.850 & 29.57/0.841 & 28.94/0.808 \\
        
        & R2R \cite{R2R} & 29.14/0.732 & 28.68/0.771 & 28.77/0.765 \\
        & NBR2NBR~\cite{neighbor2neighbor} & 30.86/0.855 & 29.54/0.843 & 29.79/0.838 \\
        & B2UNB \cite{blind2unblind} & 31.07/0.857 & 29.92/0.852 & 30.10/0.844 \\

        & DCD-Net \cite{zou2023iterative} &  31.00/0.857 & \textbf{29.99}/\textbf{0.855} & 29.99/0.843 \\

         & SST-GP \cite{yasarla2024self} & 31.39/\textbf{0.872} & \underline{29.96}/\underline{0.853} & 30.22/0.848\\

        & Noise2VST (ours)  & \textbf{31.60}/\underline{0.865} & 29.89/0.849  & \textbf{30.60}/\textbf{0.850} \\ 

        &  Noise2VST$\dagger$ (ours) & \underline{31.51}/0.862  & 29.84/0.848  & \underline{30.57}/\textbf{0.850}  \\ 
    \bottomrule
  \end{tabular}
  \caption{PSNR(dB)/SSIM denoising results in sRGB space for synthetic Poisson noise. The highest PSNR(dB)/SSIM is highlighted in \textbf{bold}, while the second is \underline{underlined}.}
  \label{tab1}
  \vspace{-0.4cm}
\end{table}

\subsection{Implementation Details}

 Our implementation is written in Python and based on the PyTorch library \cite{pytorch}. Source code is available at the following repository: \href{https://github.com/sherbret/Noise2VST}{https://github.com/sherbret/Noise2VST}. 

\paragraph{Choice of Gaussian Denoisers} The proposed method requires an off-the-shelf classic Gaussian denoiser $D$, along with a blind-spot one $\bar{D}$ that is used for VST training (see Algorithm \ref{algo1}). For the classic Gaussian denoiser, we used the pre-trained DRUNet model from \cite{drunet}, which, at the time of this writing, is the most effective convolutional neural network available for Gaussian image denoising, to the best of our knowledge. Moreover, this  denoiser  is non-blind, meaning that the noise level $\sigma$ needs to be passed as an additional argument, a characteristic  which is known to improve on generalization abilities, especially for noises that deviate significantly from the homoscedastic Gaussian assumption \cite{ffdnet, drunet}. More sophisticated transformer-based denoisers that are blind, such as \cite{restormer, xformer}, were indeed unable to effectively process real-world noise within the proposed framework in our experiments (see Supplementary Material). Those advanced models have troubles in coping with slight out-of-distribution noise, an exact variance stabilization being most of the time impossible \cite{curtiss1943transformations, vst_optimization}. As for the blind-spot denoiser, we simply adapted the previous off-the-shelf  DRUNet or the off-the-shelf FFDNet \cite{ffdnet} for the fast alternative, termed Noise2VST$\dagger$, to make them $\mathcal{J}$-invariant \cite{N2S} through \eqref{trick}, without additional training. We arbitrarily fixed the noise level of all the non-blind networks to $\sigma=25/255$, without loss of generality (see proof in Supplementary Material).

\paragraph{Training Details} 
Instead of training on the entire noisy input image $\boldsymbol{z}$, we accelerate the process by applying stochastic gradient-based optimization to overlapping $64 \times 64$ patches of $\boldsymbol{z}$. Each training iteration involves a gradient-based pass on a batch of patches that are randomly cropped from $\boldsymbol{z}$ and augmented with random vertical and horizontal flips, as well as random $90^{\circ}$ rotations. We use a batch size of 4 and employ the Adam optimizer \cite{adam} with an initial learning rate of 0.01, which is then reduced by a factor of 10 at one-third and two-thirds of the training session. Finally, the number of training iterations is set to 2000, except for raw-RGB denoising where 5000 iterations are necessary to reach convergence.

\begin{table}[t]
\scriptsize
  \centering
  \setlength{\abovecaptionskip}{0.1cm} 
  \setlength\tabcolsep{4pt}
  \begin{tabular}[b]{lccc}
    \toprule
      \multirow{2}{*}{Methods} 
      & Confocal & Confocal & Two-Photon\\
      & Fish & Mice & Mice\\
    \midrule
      Baseline, N2C \cite{unet}  & 32.79/0.905 & 38.40/0.966 & 34.02/0.925 \\
    \midrule
      BM3D  + GAT \cite{BM3D, anscombe_inverse} &  32.16/0.886 & 37.93/0.963 & 33.83/0.924 \\
        DRUNet + GAT \cite{drunet, anscombe_inverse}   &  32.18/0.879 & 38.11/0.963 & 34.01/0.925 \\

     N2V  \cite{N2V} & 32.08/0.886 & 37.49/0.960 & 33.38/0.916 \\

    PN2V  \cite{pN2V} & 32.45/0.886 & 38.24/0.960 & 33.67/0.916 \\

     ZS-N2N \cite{ZS-N2N} & 30.62/0.843 & 36.15/0.945 & 32.66/0.906 \\


      SSDN \cite{laine}  & 31.62/0.849  & 37.82/0.959 & 33.09/0.907 \\

      FBI-D \cite{FBI} & 32.22/0.885  & \underline{38.32}/0.964 & 33.95/0.908 \\

      NBR2NBR \cite{neighbor2neighbor} & 32.11/0.890 & 37.07/0.960 & 33.40/0.921 \\
      B2UNB \cite{blind2unblind} & 32.74/0.897 & \textbf{38.44}/0.964 & 34.03/0.916 \\

      Noise2VST (ours) & \textbf{32.88}/\underline{0.904} & 38.27/\textbf{0.965} & \textbf{34.06}/\underline{0.926} \\

      Noise2VST$\dagger$  (ours) & \textbf{32.88}/\textbf{0.905} & 38.26/\textbf{0.965} & \underline{34.04}/\textbf{0.927}  \\
      
    \bottomrule
  \end{tabular}
  \caption{ PSNR(dB)/SSIM denoising results on FMD dataset \cite{zhang2019poisson} in raw space.}
  \label{tab3}
\end{table}

\subsection{Synthetic Noise}

\paragraph{Datasets} We follow the same setting as in \cite{neighbor2neighbor, blind2unblind, laine} and use the three color datasets Kodak \cite{kodak}, BSD300 \cite{berkeley} and Set14 \cite{set14} for testing purposes on synthetic Poisson noise with parameter $\lambda$ in the range $[5, 50]$ (meaning that $(a, b) = (1/\lambda, 0)$ in \eqref{pg_model} with $\boldsymbol{s}_i \in [0, 1]$). 
For the sake of reliability of the average PSNR values, the datasets are repeated 10, 3, and 20 times, respectively. This allows each clean image to be corrupted by different noise instances and, in cases with varying noise parameters, by different noise levels.

\paragraph{Results} Table \ref{tab1} contains a  quantitative comparison of the results of state-of-the-art methods trained without ground truths. Note that, for SSDN \cite{laine}, only the results of the best-performing model are indicated. The baseline is established by DRUNet  with GAT \cite{anscombe_inverse}, under the assumption of the knowledge of the exact parameter $\lambda$ (oracle parameter). A first observation can be made from Table \ref{tab1}: our fully self-supervised approach performs as well as the baseline. Indeed, the performance gap between the baseline and Noise2VST is less than 0.06 dB at most for all datasets, which is not significant. This minimal impact is further supported by the additional results presented in the Supplementary Material. This is a remarkable result since our method has no prior knowledge of the noise parameters, contrary to the baseline. It indicates that the modeling and learning of the VST in our approach is close to optimal, despite the approximations inherent in the piecewise linear nature of the learned VST and the gradient-based optimization scheme. It demonstrates additionally that the traditional approach, which initially relies on an estimation of the Poisson parameters \cite{poisson_estimation, poisson_estimation2} followed by variance stabilization via GAT \cite{anscombe_inverse}, can be effectively replaced by the proposed framework. 

\input{figure_photo}

\begin{table}[t]
\scriptsize
  \centering
  \setlength{\abovecaptionskip}{0.1cm} 
  \setlength\tabcolsep{4pt}
  \begin{tabular}[b]{lccc}
    \toprule
      \multirow{2}{*}{Methods} 
      & W2S ch0 & W2S ch1 & W2S ch2\\
     & \textit{avg}1/\textit{avg}16 & \textit{avg}1/\textit{avg}16 & \textit{avg}1/\textit{avg}16 \\
    \midrule
    Raw data & 21.86/33.20 & 19.35/31.24 & 20.43/32.35 \\
      Baseline, N2C \cite{CARE}  & 34.30/41.94  & 32.11/39.09  &  34.73/40.88 \\
    \midrule
      BM3D + GAT \cite{BM3D, anscombe_inverse} &  33.33/39.31 & 31.16/38.04 & 34.50/40.63 \\
        DRUNet + GAT \cite{drunet, anscombe_inverse} &  32.54/38.88 & 30.15/37.69 & 33.75/40.03 \\
     N2V  \cite{N2V} & 34.30/38.80 & 31.80/37.81 & 34.65/40.19 \\

    PN2V  \cite{pN2V} & \quad-\quad/39.19 & \quad-\quad/38.24 & 32.48/40.49 \\
    ZS-N2N \cite{ZS-N2N} & 32.26/37.63 & 30.56/36.59 & 32.79/39.63 \\

    DivNoising \cite{divnoising} & 34.13/39.62 & 32.28/38.37 & 35.18/40.52  \\

    DecoNoising \cite{deconoising} & 34.90/39.17 & 32.31/38.33 & 35.09/40.74 \\

      Noise2VST (ours) & \underline{35.65}/\textbf{39.74} & \textbf{33.43}/\textbf{39.10} & \textbf{36.88}/\textbf{41.36} \\

      Noise2VST$\dagger$ (ours) &  \textbf{35.66}/\underline{39.70} &  \underline{33.41}/\textbf{39.10} &  \underline{36.82}/\textbf{41.36} \\
    \bottomrule
  \end{tabular}
  \caption{ PSNR denoising results (in dB) on W2S  dataset \cite{w2s} in raw space for two noise levels (\textit{avg}1 and \textit{avg}16) .}
  \label{tab2}
\end{table}

This initial observation is consistent with  the fact that our method outperforms its state-of-the-art counterparts by a significant margin on all synthetic datasets, except on BSD300 \cite{berkeley} for which a few methods \cite{blind2unblind, zou2023iterative, yasarla2024self} somehow exhibit better performance than the baseline itself. This may be attributed to the specific characteristics of BSD300, notably the prevalence of low-resolution images, while DRUNet was mainly learned on high-resolution images. Interestingly, the fast variant of Noise2VST with FFDNet, termed Noise2VST$\dagger$, performs almost on par with the heavier variant. We compare quantitatively in Figure \ref{photo2}a the image denoising results for synthetic Poisson noise with $\lambda=50$ and conclude that our method recovers the structural patterns of the glass pyramid better than the recent best B2UNB \cite{blind2unblind}, achieving almost the same quality as the baseline.

On the whole, the superiority of our method can be explained by the fact that it leverages what a pre-trained Gaussian denoiser has learned about the signal structure from a large external database composed of synthetic clean/noisy images, while all other compared methods train their denoising networks from scratch with only noisy images.

\subsection{Real-World Noise}

\paragraph{Datasets}  For real-world grayscale denoising, we consider two popular fluorescence microscopy datasets: FMD \cite{zhang2019poisson} and W2S \cite{w2s}. FMD dataset includes 12 sets of images captured through confocal, two-photon, or widefield microscopes, with each set containing 20 views and 50 different noise instances per view. For each view, the ground truth is computed by averaging the 50 noisy occurrences. In accordance with the methodology outlined in \cite{neighbor2neighbor, blind2unblind}, the 19\textsuperscript{th} 
view of three sets, namely confocal fish, confocal mice, and two-photon mice, are selected for testing purposes. 

We additionally use raw data from W2S  \cite{w2s}, acquired using a conventional fluorescence widefield. Its test set contains 40 fields of views, where each view is composed of 3 independent channels. For each view, 400 shots of independent noise instances were captured. Following the setting in \cite{w2s, pN2V, deconoising, divnoising}, the 250\textsuperscript{th} noise instance and the average of the first 16 noise instances  are used for testing and are referred to as \textit{avg1} and \textit{avg16}, respectively. The ground truth is derived by plain averaging of the 400 noise instances. 

As for image denoising in raw-RGB space, we utilize the SIDD dataset \cite{SIDD}, which comprises images captured by five smartphone cameras across 10 scenes under various lighting conditions. We employ both the SIDD validation and benchmark datasets for testing purposes.  For the benchmark dataset, the ground-truth images are
not disclosed to avoid any bias in the evaluation (the quality scores are assessed online
). It is worth noting that denoising in raw-RGB space is preferable to sRGB due to the impact of demosaicking which tends to spatially correlate the noise, thereby violating the assumption of pixel-independent noise \cite{dbsn}. 

Finally, for each real-world noisy image, the authors of the three datasets \cite{SIDD, w2s, zhang2019poisson} calculated carefully the adequate noise parameters $(a, b)$ based on a mixed Poisson-Gaussian model \eqref{pg_model} and made them available to the practitioners for GAT application \cite{anscombe_inverse}.

\paragraph{Results} We  report in Tables \ref{tab3}, \ref{tab2}, and \ref{tab4}  evaluate the PSNR/SSIM  results of state-of-the-art methods trained without specific ground truths  on the FMD, W2S  and SIDD  test sets, respectively. For raw-RGB data, the single-channel image is split into four sub-images according to the Bayer pattern and then stacked to form a four-channel image taken as input for deep learning-based methods. Since the pre-trained weights of DRUNet  and FFDNet, used within the proposed framework, are provided for single and three-channel denoising only (grayscale and color), we additionally split the four-channel image into two three-channel images (where two channels are shared) and realize denoising separately before final aggregation by averaging the shared channels. In each setting, the baseline is established by a supervised network trained on specific real-world pairs of noisy/clean images, provided by the respective datasets.

\begin{table}[t]
\scriptsize
  \centering
  \vspace{-0.2cm}
  \setlength{\abovecaptionskip}{0.1cm} 
  \setlength\tabcolsep{6pt}
  \begin{tabular}[b]{lcc}
    \toprule
      \multirow{2}{*}{Methods} 
      & SIDD & SIDD\\
      &  Benchmark & Validation\\
    \midrule
      Baseline, N2C \cite{unet} &  50.60/0.991 & 51.19/0.991 \\
    \midrule
      BM3D \cite{BM3D} + GAT \cite{anscombe_inverse}  & 48.60/0.986 & 48.92/0.986 \\
     N2V  \cite{N2V} &  48.01/0.983 & 48.55/0.984 \\
           Nr2N \cite{noisier2noise} & \textcolor{white}{---}-\textcolor{white}{---}/\textcolor{white}{---}-\textcolor{white}{---} &33.74/0.752\\

      DBSN \cite{dbsn} & 49.56/0.987 & 50.13/0.988 \\

      CycleISP  \cite{cycleisp} &  \textcolor{white}{---}-\textcolor{white}{---}/\textcolor{white}{---}-\textcolor{white}{---} &  50.45/\textcolor{white}{---}-\textcolor{white}{---}  \\

      SSDN \cite{laine} &50.28/0.989 & 50.89/0.990\\

      FBI-D \cite{FBI} &50.57/0.990 & \textcolor{white}{---}-\textcolor{white}{---}/\textcolor{white}{---}-\textcolor{white}{---}\\ 
      
      R2R \cite{R2R} &  46.70/0.978 & 47.20/0.980 \\
      NBR2NBR \cite{neighbor2neighbor} &  50.47/0.990 & 51.06/0.991 \\
      B2UNB \cite{blind2unblind} &  50.79/0.991 & 51.36/0.992 \\
       DCD-Net \cite{zou2023iterative} & \textcolor{white}{---}-\textcolor{white}{---}/\textcolor{white}{---}-\textcolor{white}{---} &51.40 0.992\\


  SST-GP \cite{yasarla2024self} & \underline{50.87}/\textbf{0.992} & \underline{51.57}/\textbf{0.992}\\

      Noise2VST (ours) & \textbf{51.07}/\underline{0.991} & \textbf{51.66}/\textbf{0.992}  \\

        Noise2VST$\dagger$ (ours) & \underline{50.87}/0.990 & 51.43/0.991  \\

    \bottomrule
  \end{tabular}
  \caption{Quantitative PSNR(dB)/SSIM denoising results on SIDD benchmark and validation datasets \cite{SIDD} in raw-RGB space.}
  \label{tab4}
\end{table}

Our approach demonstrates competitive performance compared to other methods, often achieving the best results across a majority of test sets and occasionally surpassing the baseline. The results highlight the superior capability of our method in handling complex real-world noise patterns, even though the underlying denoisers were exclusively trained for Gaussian denoising. It is worth noting that, despite sharing the same underlying philosophy, the GAT approach  generally performs poorly in real-world conditions. This can be attributed to two main factors: inaccurate estimation of noise parameters $(a, b)$ in \eqref{pg_model}; and overly simplistic nature of the Poisson-Gaussian model \eqref{pg_model} for capturing the multiple and diverse sources of degradation during data acquisition. Indeed, in addition to photon and electronic noise—traditionally modeled using a Poisson-Gaussian distribution—other perturbations may also occur during image acquisition, such as under- or over-exposure, which can also be addressed using a VST-based approach \cite{poisson_estimation}. It is perhaps not so surprising, then, that our model-free VST demonstrates strong adaptability when confronted with complex, real-world noise. These findings underline the value of a model-free VST estimation, which proves to be significantly more flexible and effective in practice. Figures \ref{photo2}b and \ref{photo2}c illustrate the superiority of our approach compared to alternative methods. Noise2VST not only reduces noise more effectively, but also recovers more texture details than its counterparts.

\subsection{Computational Efficiency}

Although Noise2VST is trained ``on the fly'' for each noisy input image, its execution time remains remarkably short, making it suitable for real-world applications 
We give in Table \ref{tab_speed} the running time of various zero-shot denoising algorithms, along with the total number of trainable parameters for each model. The tests were conducted on a Tesla V100 GPU and an Intel Xeon Gold 6240 2.60GHz CPU. It can be observed that, despite the difference in the number of trainable parameters, ZS-N2N  and Noise2VST do achieve comparable execution speeds. This is because a training pass for Noise2VST requires back-propagating the gradients through a deep neural network with frozen weights, which incurs some computational cost. Notably, the execution time of Noise2VST does not increase significantly with the size of the input image as training is performed on randomly cropped patches. Compared to the computationally intensive S2S and the fast but less effective ZS-N2N, Noise2VST presents a viable alternative for fast and efficient real-world image denoising. 


\begin{table}[t]
\scriptsize
  \centering
  \setlength{\abovecaptionskip}{0.1cm} 
  \setlength\tabcolsep{4pt}
  \begin{tabular}[b]{lcccc}
    \toprule
      Methods & S2S \cite{S2S} & ZS-N2N \cite{ZS-N2N} & Noise2VST   & Noise2VST$\dagger$  \\
    \midrule
      GPU time & 35 min.  & \textbf{20 sec.} & 50 sec.    &\textbf{20 sec.}  \\
      CPU time & 4.5 hr. & 1 min. & 5 min. & \textbf{40 sec.} \\
    \midrule
      \# parameters & 1M & 22k & 130 & 130\\
      \bottomrule
  \end{tabular}
  \caption{Running time of zero-shot denoising methods for $256\times 256$ images. Fastest is in \textbf{bold}.}
  \label{tab_speed}
\end{table}

\section{Conclusion}

We propose Noise2VST, a novel zero-shot framework for practical real-world image denoising. Our method leverages efficient off-the-shelf Gaussian denoisers trained on abundant but nonspecific synthetic clean/noisy image pairs. By deploying the recent blind-spot strategy to learn a model-free variance-stabilizing transformation for real-world noise, we demonstrate that Gaussian denoisers can achieve superior performance compared to existing methods trained without specific clean/noisy image pairs, while maintaining a reasonable computational cost. From a broader perspective, this work somehow rekindles the significance of deep learning-based Gaussian denoisers for practical uses.


\section*{Acknowledgements}
We acknowledge access to the facilities and expertise of the CIBM Center for Biomedical Imaging, a Swiss research center of excellence founded and supported by Lausanne University Hospital (CHUV), University of Lausanne (UNIL), École polytechnique fédérale de Lausanne (EPFL), University of Geneva (UNIGE), and Geneva University Hospitals (HUG).

\noindent This work was granted access to the HPC resources of IDRIS under the allocation 2024-AD011015932 made by GENCI.

{
    \small
    \bibliographystyle{ieeenat_fullname}
    \bibliography{main}
}

\end{document}